\documentclass[sigconf]{acmart}

\copyrightyear{2026}
\acmYear{2026}
\setcopyright{rightsretained}
\acmConference[epiDAMIK @ KDD '26]{8th epiDAMIK ACM SIGKDD Workshop on Data-driven Decision Making for Public and Population Health}{August 10, 2026}{Jeju Island, Republic of Korea}
\acmBooktitle{Proceedings of the 8th epiDAMIK ACM SIGKDD Workshop on Data-driven Decision Making for Public and Population Health, August 10, 2026, Jeju Island, Republic of Korea}
\acmDOI{}
\acmISBN{}

\usepackage{amsmath,amsfonts,bm}
\usepackage{booktabs}
\usepackage{array}
\usepackage{graphicx}
\usepackage{tabularx}
\usepackage{xspace}
\usepackage{placeins}

\newcommand{\method}{GeoID-PINN\xspace}
\newcommand{\Cmat}{\mathbf{C}}
\newcommand{\Czero}{\mathbf{C}_0}
\newcommand{\Chat}{\widehat{\mathbf{C}}}
\newcommand{\loss}{\mathcal{L}}

\title{GeoID-PINN: Identifiability-Aware Regional Epidemic Inference with Geographic Coupling}

\author{Weixiong Hua}
\affiliation{%
  \department{Department of Biostatistics}
  \institution{University of Michigan}
  \city{Ann Arbor}
  \state{Michigan}
  \country{USA}}

\author{Fan Bu}
\affiliation{%
  \department{Department of Biostatistics}
  \institution{University of Michigan}
  \city{Ann Arbor}
  \state{Michigan}
  \country{USA}}

\renewcommand{\shortauthors}{Hua and Bu}

\begin{document}

\begin{abstract}
Regional surveillance data reflect local transmission, reporting, seeding, and external infection pressure, which are difficult to identify separately. We introduce \method, a physics-informed neural network (PINN) for susceptible--infectious--recovered--deceased (SIRD) dynamics. The model represents spatial dependence with a row-stochastic source-composition matrix whose rows assign nonnegative source weights that sum to one. We regularize this matrix toward a spatial prior constructed from distance, adjacency, commuting, or lead--lag information.

In a four-region simulation with known truth, a compatible distance prior gives source-composition error 0.099. The error rises to 0.159 without regularization and 0.577 under a strongly misspecified prior, while trajectory fit and transmission-scale estimates remain similar. Accurate trajectories therefore do not guarantee recovery of the regional dependence structure.

We also evaluate \method retrospectively using COVID-19 data from 64 Louisiana counties. Relative to an autoregressive negative-binomial baseline, Forecast-Trained Geo-PINN reduces mean squared error (MSE) from 32,957 to 11,468 and mean absolute error (MAE) from 70.60 to 57.73. The baseline has lower negative log likelihood (NLL), 5.158 versus 5.346, indicating better distributional fit but worse point accuracy. In a controlled 15-county comparison, county adjacency reduces MSE by 6.85\% and MAE by 3.1\%. Similar performance across plausible priors supports structured regularization but not unique edge recovery. These results require prior-sensitivity and observation-model checks before interpretation.
\end{abstract}



\maketitle

\section{Introduction}
Spatial connections shape epidemic dynamics. Infection in one region can affect risk elsewhere through contact networks, commuting, travel, and other population links. Compartmental models describe disease progression \cite{kermack1927contribution,hethcote2000mathematics}, while network and metapopulation studies show that spatial structure changes synchrony, invasion timing, and regional spread \cite{keeling2005networks,eubank2004modeling,colizza2006airline,balcan2009multiscale,viboud2006synchrony,pei2018forecasting}. Several multiregion models address the related problem of specifying how epidemic activity in one location changes dynamics elsewhere. Bustamante-Casta{\~n}eda et al. couple node-level susceptible--infectious--recovered (SIR) systems through graph-Laplacian diffusion \cite{bustamante2021epidemic}. Measles studies instead construct regional influence from hierarchical or gravity-based spatial weights, or estimate latent reintroduction processes from case series across many cities \cite{finkenstadt1998empirical,xia2004measles,finkenstadt2002stochastic}. Our model shares this multiregion perspective but treats the cross-region source-composition matrix as an uncertain inverse target rather than a known movement operator.

The inverse problem is harder than forward simulation when only routine surveillance is available. Regional case curves combine local transmission, clinical removal, reporting, initial seeding, external activity, and cross-region infection pressure. These components can produce similar trajectories, so fitting the observed curve does not identify how much each component contributes.

A physics-informed neural network (PINN) represents latent state trajectories with a neural network and trains them against both observations and governing-equation residuals \cite{raissi2019pinn,karniadakis2021piml}. Epidemic PINNs can interpolate unobserved states and estimate unknown parameters, but their inverse estimates remain sensitive to the observation channels, parameterization, and regularization \cite{kharazmi2021pinn,hu2022modified,millevoi2024pinn}. This sensitivity is consistent with structural and practical identifiability results for compartmental models \cite{tuncer2018identifiability,massonis2021structural}. A regional source-composition matrix adds many more parameters and can absorb model mismatch without visibly worsening the fitted trajectories.

We therefore define a narrower target. \method does not infer unrestricted movement or claim to recover a causal transmission network. For each recipient region, it represents infection pressure as a weighted average of infectious prevalence in the source regions. The weights form a row-stochastic \emph{source-composition matrix}. Its interpretation is conditional on the SIRD model, observation process, parameter constraints, and spatial prior.

The two evaluations answer different questions. The four-region synthetic benchmark has a known source-composition matrix and tests recovery under changes to the prior and observation process. The 64-county Louisiana study has no true edge map. It instead evaluates retrospective forecast accuracy, baseline performance, and robustness across practical priors. Good forecasts in Louisiana do not validate a unique matrix.

\method adds three safeguards to a standard epidemic PINN. It constrains the SIRD states to remain nonnegative and preserve each regional population total. It places a row-stochastic source-composition matrix inside the force of infection and regularizes that matrix toward an empirical spatial prior. It also separates observation fit, mechanistic consistency, and matrix recovery during evaluation. The paper uses these distinctions throughout.

\section{Problem Setup and Identifiability Risk}
\subsection{Regional SIRD model and source composition}
Consider $P$ regions. Index $p\in\{1,\ldots,P\}$ denotes a recipient region, and $q\in\{1,\ldots,P\}$ denotes a source region. At time $t$, the states $S_p(t)$, $I_p(t)$, $R_p(t)$, and $D_p(t)$ are susceptible, infectious, recovered, and cumulative deaths in region $p$. The symbol $D$ has the same meaning in both studies. The fixed population is $N_p=S_p(t)+I_p(t)+R_p(t)+D_p(t)$. The living population is $N^{\mathrm{live}}_p(t)=S_p(t)+I_p(t)+R_p(t)$. Recovery and mortality rates are $\gamma_p\geq 0$ and $\mu_p\geq 0$. A dot denotes differentiation with respect to time.

The force of infection $\lambda_p(t)$ is the instantaneous per-susceptible infection hazard in recipient $p$. Let $\alpha\geq 0$ be the global transmission scale and let $\Cmat=[C_{pq}]$ be a nonnegative $P\times P$ source-composition matrix, where $C_{pq}$ is the source-$q$ weight for recipient $p$. Then
\begin{equation}
\lambda_p(t)=\alpha\sum_{q=1}^{P} C_{pq}
\frac{I_q(t)}{N^{\mathrm{live}}_q(t)},
\qquad C_{pq}\geq 0,\quad \sum_{q=1}^{P}C_{pq}=1,
\label{eq:foi}
\end{equation}
The matrix is \emph{row-stochastic}: the entries in each recipient row sum to one. Thus $C_{pq}$ is the share of infection pressure in recipient $p$ assigned to source $q$, and the diagonal entry $C_{pp}$ is the local share. The four SIRD balance equations are
\begin{subequations}\label{eq:sird}
\begin{align}
\dot S_p(t) &= -S_p(t)\lambda_p(t), \label{eq:sird-s}\\
\dot I_p(t) &= S_p(t)\lambda_p(t)-\{\gamma_p+\mu_p\}I_p(t), \label{eq:sird-i}\\
\dot R_p(t) &= \gamma_p I_p(t), \label{eq:sird-r}\\
\dot D_p(t) &= \mu_p I_p(t). \label{eq:sird-d}
\end{align}
\end{subequations}
This infection operator is compatible with structured compartmental models \cite{diekmann2010nextgeneration}. Its rows describe source composition, not the number of people moving between regions. Thus $\Cmat$ is neither a mobility matrix nor an edge-level causal quantity.

For comparison, a conventional metapopulation model adds movement to each living compartment. Let $X_p(t)$ denote $S_p(t)$, $I_p(t)$, or $R_p(t)$, and let $X(t)$ collect that compartment across all regions. The function $F_p\{X(t);\boldsymbol\eta_p\}$ is the within-region disease operator with local parameter vector $\boldsymbol\eta_p$. The rate $J_{pq}(t)\geq0$ is the per-capita flow from region $p$ to region $q$, so $J_{qp}(t)$ is flow from $q$ into $p$. Such a model has
\begin{equation}
\dot X_p(t)=F_p\{X(t);\boldsymbol\eta_p\}
+\sum_{q\neq p}J_{qp}(t)X_q(t)
-X_p(t)\sum_{q\neq p}J_{pq}(t).
\label{eq:movement}
\end{equation}
Equation~\eqref{eq:movement} is useful when flows are observed or externally constrained. Estimating every $J_{pq}(t)$ from epidemic curves alone is much less determined. Movement can trade off against local transmission, removal, unobserved seeding, reporting, and neural-state flexibility. \method therefore does not fit Eq.~\eqref{eq:movement}. Source composition appears only in the force of infection in Eq.~\eqref{eq:foi}, which supports a narrower interpretation.

\subsection{Why the matrix is weakly identified}
Let $v_q(t)=I_q(t)/N^{\mathrm{live}}_q(t)$ be infectious prevalence in source $q$. The weighted prevalence for recipient $p$ is $h_p(t)=\sum_{q=1}^{P} C_{pq}v_q(t)$. Equation~\eqref{eq:foi} depends on row $p$ of $\Cmat$ only through $h_p(t)$. Let $\Delta C_{p\cdot}$ perturb that entire row; the centered dot means all columns. When regional prevalence curves are correlated, many perturbations satisfy both $\sum_q\Delta C_{pq}=0$ and $\sum_q\Delta C_{pq}v_q(t)\approx 0$. They produce nearly the same infection pressure even though the individual weights differ; here $\approx$ denotes approximate equality.

Row normalization clarifies the roles of $\alpha$ and $\Cmat$, but it does not remove the ambiguity. Overall epidemic growth may identify the scalar magnitude $\alpha$ more strongly than any individual entry of $\Cmat$. A spatial prior can then select one matrix from a weakly distinguished set while leaving the fitted trajectory almost unchanged. Small state or incidence residuals are therefore not evidence of edge recovery.

Other model components create further ambiguity. Reporting intensity changes the apparent prevalence of a source region. Reporting delays can shift apparent lead--lag relationships. Removal rates, initial seeding, and flexible neural states can also absorb local mismatch. The freeze--unfreeze optimizer may add path dependence because the state network is first fit with the source-composition matrix fixed. That path is not, however, the main explanation for prior sensitivity. The broader problem is practical non-identifiability of the model under the available observations.

\section{GeoID-PINN Method}
A standard PINN represents latent state trajectories with a neural network and balances data fit against governing-equation residuals. Unknown model parameters can be optimized with the network weights. \method keeps this structure and adds safeguards for regional inference. The neural SIRD states remain nonnegative and sum to each regional population. A row-stochastic source-composition matrix enters the force of infection, and an empirical spatial prior regularizes that matrix.

\begin{figure*}[t]
\centering
\includegraphics[width=0.98\textwidth]{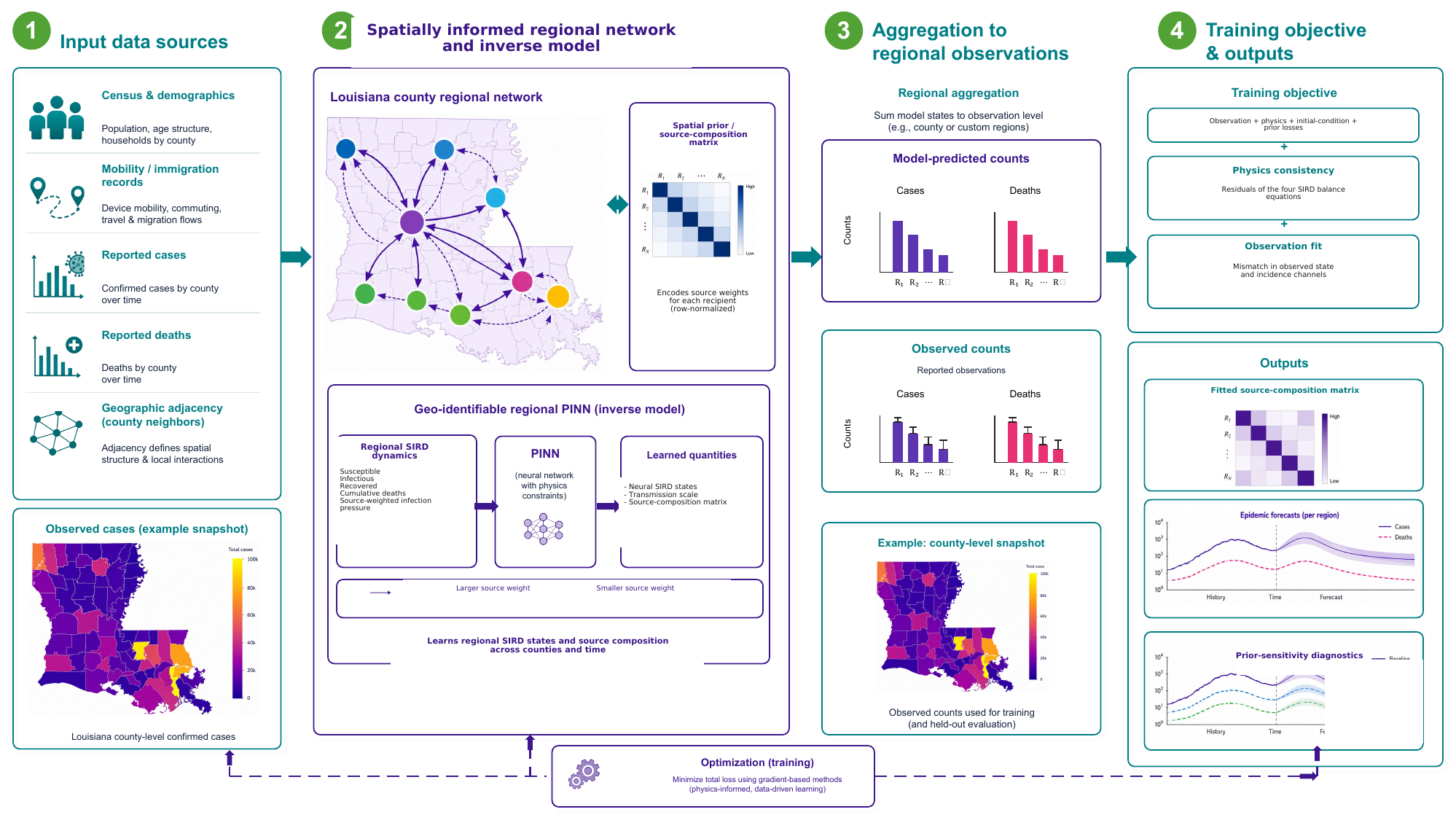}
\caption{Conceptual GeoID-PINN framework. The diagram summarizes candidate demographic, mobility, and adjacency inputs; a regional PINN; aggregation to observed counts; and training outputs. Its model and objective labels use the SIRD states and loss components defined in the text. The matrix $\Cmat$ is conditional source composition, not literal mobility or a causal transmission network.}
\Description{A four-stage diagram shows regional data sources, a population- and mobility-informed regional PINN, aggregation to observed counts, and training objectives and outputs. The manuscript interprets the displayed regional matrix as conditional source composition rather than literal mobility or causal transmission.}
\label{fig:idea}
\end{figure*}

Figure~\ref{fig:idea} gives a conceptual overview of the inputs, regional model, observation aggregation, training objective, and outputs. The diagram includes mobility and immigration as possible information sources. In the fitted model, however, $\Cmat$ is interpreted only as conditional source composition, not literal mobility or causal transmission. Labels of the form R1, R2, and so on inside the schematic are generic region identifiers; they are not the recovered state $R_p(t)$.

\subsection{Constrained neural states and source composition}
Let $u_{\boldsymbol\phi}(t)$ denote the state network, with weights and biases $\boldsymbol\phi$. For region $p$ and compartment $c\in\{S,I,R,D\}$, the network outputs an unconstrained logit $a_{pc}(t)$. Let $\exp$ denote the exponential function and let $c'$ be a dummy compartment index. We convert the logit to the constrained state $X_{pc}(t)$ by
\begin{equation}
X_{pc}(t)=N_p\frac{\exp\{a_{pc}(t)\}}
{\sum_{c'\in\{S,I,R,D\}}\exp\{a_{pc'}(t)\}},
\label{eq:state-softmax}
\end{equation}
The mappings are $X_{pS}=S_p$, $X_{pI}=I_p$, $X_{pR}=R_p$, and $X_{pD}=D_p$. This construction keeps every state nonnegative and makes the four states sum to $N_p$.

Let $\mathbf B=[B_{pq}]$ be the unconstrained coupling logits. The structural mask $\mathbf A=[A_{pq}]$ sets $A_{pq}=1$ when source $q$ is allowed for recipient $p$ and $A_{pq}=0$ otherwise. Let $q'$ be a dummy source index. If each recipient has at least one allowed source, the masked row-wise softmax is
\begin{equation}
C_{pq}=\frac{A_{pq}\exp(B_{pq})}
{\sum_{q'=1}^{P}A_{pq'}\exp(B_{pq'})},
\label{eq:coupling-softmax}
\end{equation} The unmasked model sets $A_{pq}=1$ for every pair. Depending on the experiment, $\gamma_p$ and $\mu_p$ are fixed or learned within $[1/28,1/3]$ and $[10^{-5},0.05]$ through sigmoid constraints. Reporting factors remain fixed in all reported experiments. Learning them jointly with $\Cmat$ from incidence alone would add another source of non-identifiability.

\subsection{Observations, normalization, and complete loss}
Let $t_k$, for $k\in\{0,\ldots,K-1\}$ and $K\geq2$, denote the model times. A hat denotes a model prediction. For channel $c\in\{I,R,D\}$, $Y^c_{pk}$ is the observed value and $\widehat Y^c_{pk}=X_{pc}(t_k)$ is the neural prediction. The mask $M^c_{pk}\in\{0,1\}$ equals one when the observation is available. The superscript $c$ labels a channel, not a power. Let $Z_{pk}$ be observed incidence in region $p$ over $[t_k,t_{k+1}]$, with mask $M^Z_{pk}$. The model-implied latent incidence is
\begin{equation}
\widehat Z_{pk}=S_p(t_k)-S_p(t_{k+1}),
\label{eq:incidence}
\end{equation}
The fixed reporting factor $\pi_p>0$ maps latent incidence to the predicted reported count $\pi_p\widehat Z_{pk}$.

For generic arrays $\widehat Y$, $Y$, and $M$, let $\Omega_M=\{(p,k):M_{pk}=1\}$ be the observed index set. Its size is $|\Omega_M|$. For a fixed positive scale $s$, the normalized mean squared error (NMSE) is
\begin{equation}
\operatorname{NMSE}(\widehat Y,Y,M;s)=
\frac{1}{|\Omega_M|}\sum_{(p,k)\in\Omega_M}
\left(\frac{\widehat Y_{pk}-Y_{pk}}{s}\right)^2.
\label{eq:nmse}
\end{equation}
The positive constants $s_I$, $s_R$, $s_D$, and $s_Z$ scale the infectious, recovered, deceased, and incidence channels. We calculate them once from the observed data before optimization and do not learn them. The symbol $s_c$ denotes the scale for channel $c$. Let $\pi\widehat Z$ denote the array with entries $\pi_p\widehat Z_{pk}$. The state, incidence, and combined data losses are
\begin{align}
\loss_{\mathrm{state}} &=
\sum_{c\in\{I,R,D\}}\operatorname{NMSE}
(\widehat Y^c,Y^c,M^c;s_c), \label{eq:state-loss}\\
\loss_{\mathrm{inc}} &=
\operatorname{NMSE}(\pi\widehat Z,Z,M^Z;s_Z), \label{eq:inc-loss}\\
\loss_{\mathrm{data}} &= \loss_{\mathrm{state}}+20\loss_{\mathrm{inc}},
\label{eq:data-loss}
\end{align}
Each state channel has weight 1, and incidence has weight 20.

For $k\in\{0,\ldots,K-2\}$, let $\Delta t_k=t_{k+1}-t_k$. For compartment $c\in\{S,I,R,D\}$, define $\delta_tX^c_{pk}=\{X_{pc}(t_{k+1})-X_{pc}(t_k)\}/\Delta t_k$. We use $\delta_tS_{pk}$, $\delta_tI_{pk}$, $\delta_tR_{pk}$, and $\delta_tD_{pk}$ as shorthand. The four SIRD residuals are
\begin{subequations}\label{eq:residuals}
\begin{align}
\varepsilon^S_{pk}&=\delta_tS_{pk}+S_p(t_k)\lambda_p(t_k),\\
\varepsilon^I_{pk}&=\delta_tI_{pk}-S_p(t_k)\lambda_p(t_k)
 +(\gamma_p+\mu_p)I_p(t_k),\\
\varepsilon^R_{pk}&=\delta_tR_{pk}-\gamma_pI_p(t_k),\\
\varepsilon^D_{pk}&=\delta_tD_{pk}-\mu_pI_p(t_k).
\end{align}
\end{subequations}
For each compartment $c$, $s_{\dot c}>0$ is a fixed derivative scale; the dot indicates a time derivative. The normalized physics loss is
\begin{equation}
\loss_{\mathrm{phys}}=
\frac{1}{4P(K-1)}\sum_{p=1}^{P}\sum_{k=0}^{K-2}
\sum_{c\in\{S,I,R,D\}}
\left(\frac{\varepsilon^c_{pk}}{s_{\dot c}}\right)^2.
\label{eq:phys-loss}
\end{equation}
Let $x^0_{pc}$ be the supplied initial state for region $p$ and compartment $c$. The initial-condition (IC) loss averages the raw squared error over all $4P$ initial states:
\begin{equation}
\loss_{\mathrm{IC}}=\frac{1}{4P}\sum_{p=1}^{P}
\sum_{c\in\{S,I,R,D\}}
\left\{X_{pc}(t_0)-x^0_{pc}\right\}^2.
\label{eq:ic-loss}
\end{equation}

Let $\Czero=[(C_0)_{pq}]$ be the row-stochastic prior matrix. Its regularization strength is $\lambda_C\geq0$, and the fixed coupling scale is $s_C=0.10$. The prior loss is
\begin{equation}
\loss_{\mathrm{prior}}=
\lambda_C\frac{1}{P^2}\sum_{p=1}^{P}\sum_{q=1}^{P}
\left\{\frac{C_{pq}-(C_0)_{pq}}{s_C}\right\}^2.
\label{eq:prior-loss}
\end{equation}
The synthetic experiments use the following core objective. The Louisiana specifications retain the same component losses:
\begin{equation}
\loss=\loss_{\mathrm{state}}+20\loss_{\mathrm{inc}}
+\loss_{\mathrm{phys}}+10\loss_{\mathrm{IC}}
+\loss_{\mathrm{prior}}.
\label{eq:total-loss}
\end{equation}
The fixed weights are 1 for each state channel, 20 for incidence, 1 for the physics loss, and 10 for initial conditions. The value of $\lambda_C$ varies by scenario. In Louisiana, county balancing changes each county's contribution within the data term, as described in Section~\ref{sec:louisiana-design}. A separate code branch adds a reporting-prior term with coefficient 0.10 only when $\pi_p$ is learned. Reporting is fixed here, so that branch is inactive and does not appear in Eq.~\eqref{eq:total-loss}.

\subsection{Optimization and spatial priors}
Training proceeds in two stages. We first fix the observation masks and scales, choose the structural mask $\mathbf A$, and initialize $\mathbf B$ so that Eq.~\eqref{eq:coupling-softmax} reproduces $\Czero$ on the allowed support. During warmup, $\mathbf B$ is frozen while $\boldsymbol\phi$, $\alpha$, and any learnable clinical rates fit the trajectory scale. We then unfreeze $\mathbf B$ and optimize all active parameters together. One epoch is one pass over the fixed model grid. The selected synthetic run uses 800 adaptive moment estimation (Adam) epochs and freezes $\mathbf B$ for the first 200. The grid experiments use 400 epochs and freeze it for the first 100. The synthetic learning rate is $3\times10^{-3}$. This schedule reduces optimization imbalance in PINNs \cite{wang2021gradient,hu2022modified,millevoi2024pinn}, but it cannot resolve non-identifiability.

We evaluate each fit with three separate diagnostics. Channel-specific losses measure observation fit, the four SIRD residuals measure mechanistic consistency, and prior perturbations assess the source-composition matrix. When the true matrix is known, we also report matrix error. A small physics residual does not imply good observation fit or correct edge recovery.

For the synthetic priors, $d_{pq}\geq0$ is the Euclidean distance between regions $p$ and $q$, and $\ell>0$ controls the distance decay. For $p\neq q$, we use the distance and anti-distance kernels
\begin{align}
W^{\mathrm{dist}}_{pq}&=\exp(-d_{pq}/\ell),\\
W^{\mathrm{anti}}_{pq}&=\exp(d_{pq}/\ell),
\label{eq:distance-weight}
\end{align}
with $W_{pp}=0$. The superscripts ``dist'' and ``anti'' distinguish the two constructions, and $W_{pq}$ denotes the chosen kernel. The \emph{diagonal mass} $\rho\in[0,1]$ is the prescribed local-source share $(C_0)_{pp}$. Let $q'$ be a dummy source index. We distribute the remaining mass across nonlocal sources and normalize each recipient row:
\begin{equation}
(C_0)_{pp}=\rho,\qquad
(C_0)_{pq}=(1-\rho)\frac{W_{pq}}
{\sum_{q'\neq p}W_{pq'}},\quad p\neq q,
\label{eq:prior-construction}
\end{equation} Equation~\eqref{eq:prior-construction} fully specifies the distance-based $\Czero$: the kernel, diagonal mass, length scale, and row normalization are all explicit.

The same construction applies to other geographic scores. Let $G_{pq}\geq0$ be an off-diagonal score with $G_{pp}=0$, and let $\rho_p\in[0,1]$ be the desired local mass for recipient $p$. Replacing $W_{pq}$ with $G_{pq}$ and $\rho$ with $\rho_p$ converts adjacency, distance, commuting, or residual lead--lag information into a row-stochastic prior. Rows always index recipients, and columns index sources. If every off-diagonal score in a row is zero, the implementation must use a prespecified identity or fallback row. Louisiana commuting priors use directed scores from the U.S. Census Longitudinal Employer-Household Dynamics Origin-Destination Employment Statistics (LODES). The final data citation should report the release, access date, and preprocessing convention.

\begin{table}[tbp]
\caption{Candidate constructions for $\Czero$. Each provides regularization or initialization; none is verified causal transmission.}
\label{tab:practical-priors}
\centering
\small
\begin{tabularx}{\columnwidth}{@{}>{\raggedright\arraybackslash}p{0.20\columnwidth}>{\raggedright\arraybackslash}p{0.28\columnwidth}X@{}}
\toprule
Source & Off-diagonal score $G_{pq}$ & Main interpretation constraint \\
\midrule
Adjacency & Shared-boundary indicator & Encodes topology, not interaction magnitude. \\
Distance & $\exp(-d_{pq}/\ell)$ & Depends on the chosen length scale. \\
LODES commuting & Nonnegative directed commuting score & Measured travel is not equivalent to transmission. \\
Residual lead--lag & Nonnegative association after local fitting & May reflect reporting delay or common forcing. \\
Identity & Set $\Czero=\mathbf I_P$, the $P\times P$ identity matrix & Local-only; no cross-region mass. \\
\bottomrule
\end{tabularx}
\end{table}

In Louisiana, the \emph{Data-Inferred Transmission Network} uses adaptive prewhitened cross-correlation (PWCCF). This residual lead--lag score is computed after filtering local serial dependence. We compare it with adjacency, distance, commuting, placebo, identity, and unregularized alternatives. Similar forecast performance across these choices supports structured regularization, not a uniquely recovered edge map.

\section{Experimental Design}
\begin{table*}[tbp]
\caption{Roles of the two evaluations. The synthetic study tests recovery against known coupling; the Louisiana study tests retrospective forecast accuracy and prior robustness.}
\label{tab:study-roles}
\centering
\small
\begin{tabularx}{\textwidth}{@{}>{\raggedright\arraybackslash}p{0.16\textwidth}>{\raggedright\arraybackslash}p{0.19\textwidth}X X@{}}
\toprule
Study & Scale and repetition & Primary evidence & Claim not supported \\
\midrule
Four-region synthetic & 4 regions, 90 daily times, one seed per reported grid & Coupling error against the known true matrix; prior and observation stress & Large-scale utility, multistart uncertainty, or real-world network recovery \\
Louisiana retrospective & 64 counties; about 44 weeks; four origins; three neural seeds & Baselines, point and likelihood metrics, controlled geography ablation, practical-prior robustness & A unique transmission network, independent external validation, or deployable real-time forecasting \\
\bottomrule
\end{tabularx}
\end{table*}

\subsection{Four-region known-truth study}
The synthetic benchmark has four regions and 90 daily time points. A superscript $*$ denotes data-generating truth, and a hat denotes an estimate. Thus $\alpha^*$ and $\Cmat^*$ are the true transmission scale and source-composition matrix, while $\hat\alpha$ and $\Chat$ are their fitted counterparts. The region coordinates are $(0,0)$, $(1,0.2)$, $(0.4,1.2)$, and $(1.4,1.1)$. Populations are $(1.00,0.78,0.62,0.55)$, and initial infectious states are $(0.0040,0.0015,0.0008,0.0004)$. The true transmission scale is $\alpha^*=0.42$. Recovery rates are $(0.100,0.095,0.105,0.098)$, and mortality rates are $(0.0030,0.0035,0.0027,0.0032)$. The true source-composition matrix is
\begin{equation}
\Cmat^*=\begin{bmatrix}
0.80&0.12&0.06&0.02\\
0.10&0.78&0.08&0.04\\
0.05&0.12&0.78&0.05\\
0.02&0.07&0.11&0.80
\end{bmatrix}.
\label{eq:true-c}
\end{equation}
The state network has two hidden layers of width 32 with hyperbolic tangent ($\tanh$) activations. Under clean full-state observations, the prior grid changes only the kernel, diagonal mass, length scale, and $\lambda_C$. ``No prior'' sets $\lambda_C=0$. Its distance-based $\Czero$ initializes the logits but adds no penalty. A second grid fixes the medium distance prior and varies the observation channel: clean full, noisy full, weekly sparse, reporting-biased, or cases-only proxy. Each reported grid uses one seed, so multistart confidence intervals are unavailable. For a matrix $\mathbf Q=[Q_{pq}]$, the Frobenius norm is $\lVert\mathbf Q\rVert_F=(\sum_{p,q}Q_{pq}^2)^{1/2}$. We report coupling error as $\lVert\Chat-\Cmat^*\rVert_F$.

\subsection{Louisiana retrospective forecasting case}
\label{sec:louisiana-design}
The Louisiana case uses weekly county COVID-19 surveillance data from March 5 through December 31, 2020. The series were obtained through the COVID Neighborhood Project (CONEP), which assembled spatially referenced case data from state health departments \cite{noppert2024state}. The analysis includes 64 counties and about 44 weekly observations. A \emph{forecast origin} is the last date treated as observed before a six-week test window. The four origins are July 23, September 3, October 15, and November 26, 2020. Each neural specification is run with three random seeds. These repeated origins and seeds come from one data source; they are not independent external validation.

The \emph{Forecast-Trained Geo-PINN} has three main components. First, an observed external infection-pressure input represents activity outside the modeled county network. Second, the training objective combines raw-count errors with county-normalized errors, preventing the largest counties from dominating the fit. Third, \emph{masked-origin training} removes selected post-origin observations from the observation loss and creates forecast-like tasks. The network predicts states directly at forecast times. It does not recursively roll the SIRD equations forward and does not use a local autoregressive correction. Because the external input is observed during the test window, this is a retrospective, \emph{oracle-input} evaluation. An operational system would have to forecast that input separately.

The 64-county comparison includes a Local-Only PINN and an Autoregressive Negative-Binomial Baseline. The Local-Only PINN removes both cross-county coupling and the external infection-pressure channel, so it compares two bundled systems. A separate analysis on a 15-county high-case subset changes only the coupling. The No County-Network PINN uses identity coupling, while the Neighbor-Network Geo-PINN uses county adjacency. All other inputs and training steps are retained. This pair isolates the contribution of explicit geography more cleanly.

For record $j\in\{1,\ldots,n\}$, let $y_j$ be the observed count and $\widehat m_j$ the predicted mean. The total number of county--origin--horizon records is $n$. Mean squared error (MSE) and mean absolute error (MAE) are defined below; the vertical bars in the MAE formula denote absolute value:
\begin{equation}
\mathrm{MSE}=\frac{1}{n}\sum_{j=1}^{n}(y_j-\widehat m_j)^2,
\qquad
\mathrm{MAE}=\frac{1}{n}\sum_{j=1}^{n}|y_j-\widehat m_j|.
\label{eq:point-metrics}
\end{equation}
For a baseline model and a proposed model, the subscripts ``baseline'' and ``proposed'' identify those two models. Positive relative gains mean that the proposed model has lower point error:
\begin{equation}
\begin{aligned}
\mathrm{Relative\ MSE\ Gain}
&=\frac{\mathrm{MSE}_{\mathrm{baseline}}-\mathrm{MSE}_{\mathrm{proposed}}}
{\mathrm{MSE}_{\mathrm{baseline}}},\\
\mathrm{Relative\ MAE\ Gain}
&=\frac{\mathrm{MAE}_{\mathrm{baseline}}-\mathrm{MAE}_{\mathrm{proposed}}}
{\mathrm{MAE}_{\mathrm{baseline}}}.
\end{aligned}
\label{eq:relative-gains}
\end{equation}
For a generic count $y$ and predictive mean $m$, let $p_{\theta_{\mathrm{NB}}}(y\mid m)$ be the negative-binomial probability of $y$ conditional on $m$; the vertical bar denotes conditioning. The subscript ``NB'' denotes negative binomial, $\theta_{\mathrm{NB}}=2$ is the fixed dispersion, and $\log$ is the natural logarithm. Mean negative log likelihood (NLL) is
\begin{equation}
\mathrm{NLL}=-\frac{1}{n}\sum_{j=1}^{n}
\log p_{\theta_{\mathrm{NB}}}(y_j\mid\widehat m_j),
\label{eq:nll}
\end{equation}
The fixed dispersion is a downstream experimental compromise rather than the optimum of an earlier grid. Lower MSE, MAE, and NLL are better. MSE and MAE assess the predicted mean, whereas NLL also evaluates the assumed count distribution. A model can therefore improve point accuracy yet have worse NLL. The analysis uses aggregate county-week counts and no individual-level records.

\section{Results}
\subsection{Synthetic prior sensitivity}
Table~\ref{tab:synthetic-priors} reports the exact prior settings and the clean-observation results. Distance weak and Distance medium are closest to the true matrix, with Frobenius errors 0.094 and 0.099. Without a penalty, the error increases to 0.159. Deliberately mismatched composition or diagonal mass raises it to 0.341--0.577. The normalized data losses, however, stay within 0.00322--0.00396.

\begin{table*}[tbp]
\caption{Four-region synthetic prior definitions and clean-observation outcomes. Lower coupling error and data loss are better. For No prior, $\lambda_C=0$ and $\Czero$ is initialization only.}
\label{tab:synthetic-priors}
\centering
\small
\begin{tabular}{@{}llcccrrr@{}}
\toprule
Prior & Kernel & Diagonal $\rho$ & Length $\ell$ & $\lambda_C$ & $\hat\alpha$ & $\lVert\Chat-\Cmat^*\rVert_F$ & Data loss \\
\midrule
Distance weak   & Distance       & 0.80 & 1.0 & 0.03 & 0.463 & 0.094 & 0.00325 \\
Distance medium & Distance       & 0.80 & 1.0 & 0.20 & 0.463 & 0.099 & 0.00322 \\
No prior        & Distance init.  & 0.80 & 1.0 & 0.00 & 0.463 & 0.159 & 0.00332 \\
Too diagonal    & Distance       & 0.93 & 1.0 & 0.20 & 0.462 & 0.341 & 0.00345 \\
Too mixed       & Distance       & 0.60 & 1.0 & 0.20 & 0.465 & 0.449 & 0.00327 \\
Wrong strong    & Anti-distance  & 0.55 & 2.5 & 1.00 & 0.468 & 0.577 & 0.00396 \\
\bottomrule
\end{tabular}
\end{table*}

The transmission-scale estimate is much more stable: $\hat\alpha$ ranges from 0.462 to 0.468. Equation~\eqref{eq:foi} explains the difference. Row normalization assigns overall magnitude to $\alpha$, while correlated regional prevalence allows many source compositions to produce nearly the same weighted prevalence. The data thus constrain epidemic growth more strongly than individual weights. The prior selects among those similar rows. Freeze--unfreeze training may add path dependence, but it does not by itself explain the sixfold range in matrix error.

\subsection{Synthetic observation stress}
With the medium distance prior fixed, clean, noisy, and weekly sparse observations give coupling errors of 0.098--0.099 and data losses below 0.0047 (Table~\ref{tab:observation-stress}). Reporting bias and the cases-only proxy raise $\hat\alpha$ to 0.537 and 0.545. Their data losses increase sharply to 0.746 and 0.984. The coupling errors remain near 0.099 because the fitted matrices stay close to the prior, not because the misspecified observations confirm the matrix.

\begin{table}[tbp]
\caption{Four-region synthetic observation stress under the medium distance prior. Lower errors and losses are better; large data loss flags mismatch.}
\label{tab:observation-stress}
\centering
\small
\begin{tabular}{@{}lrrrr@{}}
\toprule
Observation & $\hat\alpha$ & $\lVert\Chat-\Cmat^*\rVert_F$ & Data loss & Physics loss \\
\midrule
Clean full       & 0.463 & 0.099 & 0.00322 & 0.00288 \\
Noisy full       & 0.465 & 0.099 & 0.00461 & 0.00293 \\
Weekly sparse    & 0.468 & 0.098 & 0.00403 & 0.00344 \\
Reporting bias   & 0.537 & 0.098 & 0.74570 & 0.00055 \\
Cases-only proxy & 0.545 & 0.099 & 0.98395 & 0.00239 \\
\bottomrule
\end{tabular}
\end{table}

\begin{figure*}[tbp]
\centering
\includegraphics[width=0.91\textwidth]{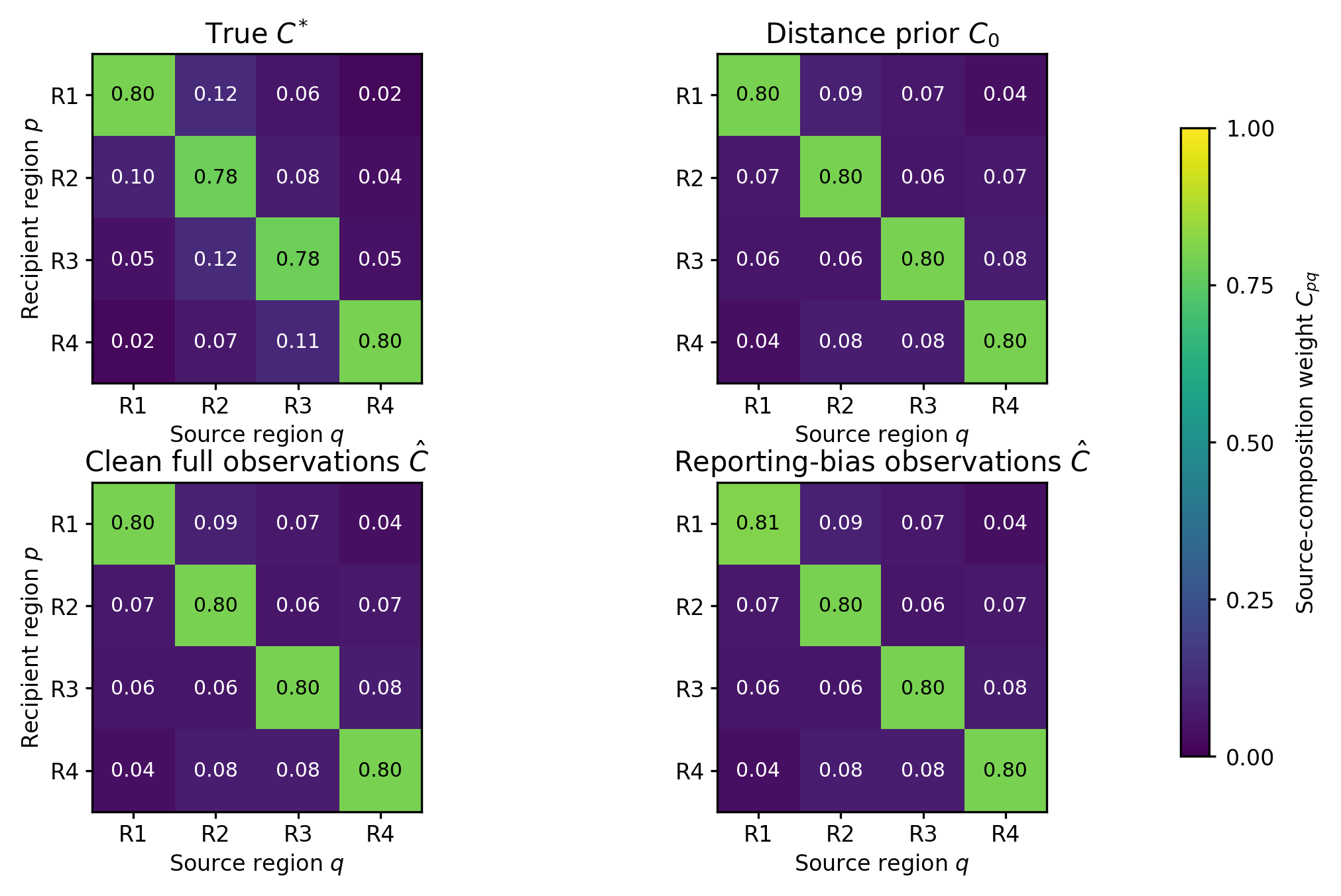}
\caption{Synthetic observation-stress matrices on a shared 0--1 scale. The reporting-biased estimate remains close to the distance prior, but its large data loss in Table~\ref{tab:observation-stress} shows that the prior dominates under observation-model mismatch. This is not successful recovery. Values are shown to two decimals; R1--R4 denote synthetic regions 1--4.}
\Description{Four equally sized four-by-four heatmaps compare the true coupling matrix, distance prior, clean-observation estimate, and reporting-bias estimate. Every panel uses the same zero-to-one color scale and one labeled colorbar. Rows are recipient regions and columns are source regions.}
\label{fig:observation-heatmaps}
\end{figure*}

Figure~\ref{fig:observation-heatmaps} shows why the diagnostics must be read together. A matrix can look plausible and remain close to $\Czero$ even when the observation fit is poor. Heatmaps alone are not enough; they must be paired with observation-channel residuals.

\subsection{Louisiana 64-county progression and baseline tradeoff}
Table~\ref{tab:all-models} shows the model progression and principal baselines across all 64 counties. The intermediate rows record successive specifications. They are not one-factor ablations because several changes accumulate across the sequence.

\begin{table}[tbp]
\caption{Louisiana retrospective model progression across all 64 counties; lower MSE, MAE, and NLL are better. Bold marks the lowest value in each metric column.}
\label{tab:all-models}
\centering
\small
\begin{tabularx}{\columnwidth}{@{}>{\raggedright\arraybackslash}Xrrr@{}}
\toprule
Model & MSE & MAE & NLL \\
\midrule
Local-Only PINN & 25,297 & 81.77 & 5.671 \\
Closed-System Geo-PINN & 27,568 & 79.41 & 5.623 \\
Frozen-Input Geo-PINN & 21,786 & 69.17 & 5.653 \\
Frozen-Input Balanced Geo-PINN & 21,318 & 68.80 & 5.660 \\
Observed-Input Geo-PINN & 14,492 & 62.99 & 5.349 \\
Observed-Input Balanced Geo-PINN & 13,970 & 62.44 & 5.352 \\
Forecast-Trained Geo-PINN & \textbf{11,468} & \textbf{57.73} & 5.346 \\
Autoregressive Negative-Binomial Baseline & 32,957 & 70.60 & \textbf{5.158} \\
\bottomrule
\end{tabularx}
\end{table}

Forecast-Trained Geo-PINN lowers MSE from 32,957 to 11,468 and MAE from 70.60 to 57.73 relative to the Autoregressive Negative-Binomial Baseline. By Eq.~\eqref{eq:relative-gains}, these are gains of 65.2\% in MSE and 18.2\% in MAE. The baseline has lower NLL, 5.158 versus 5.346. MSE and MAE score the predictive mean. NLL instead scores the probability assigned to each observed count under the assumed negative-binomial distribution, so it also reflects the distribution's variance. The baseline is trained with a negative-binomial likelihood. Geo-PINN is trained with observation, physics, initial-condition, and prior losses; its NLL is computed afterward with fixed $\theta_{\mathrm{NB}}=2$. The comparison therefore favors Geo-PINN for point accuracy and the baseline for distributional fit.

Compared with the Local-Only PINN, the final model lowers MSE from 25,297 to 11,468, a 54.7\% gain. MAE falls from 81.77 to 57.73, a 29.4\% gain, and NLL falls from 5.671 to 5.346. This comparison changes both county coupling and the external infection-pressure channel, so it does not isolate the effect of geography.

\begin{figure*}[!t]
\centering
\includegraphics[width=0.90\textwidth]{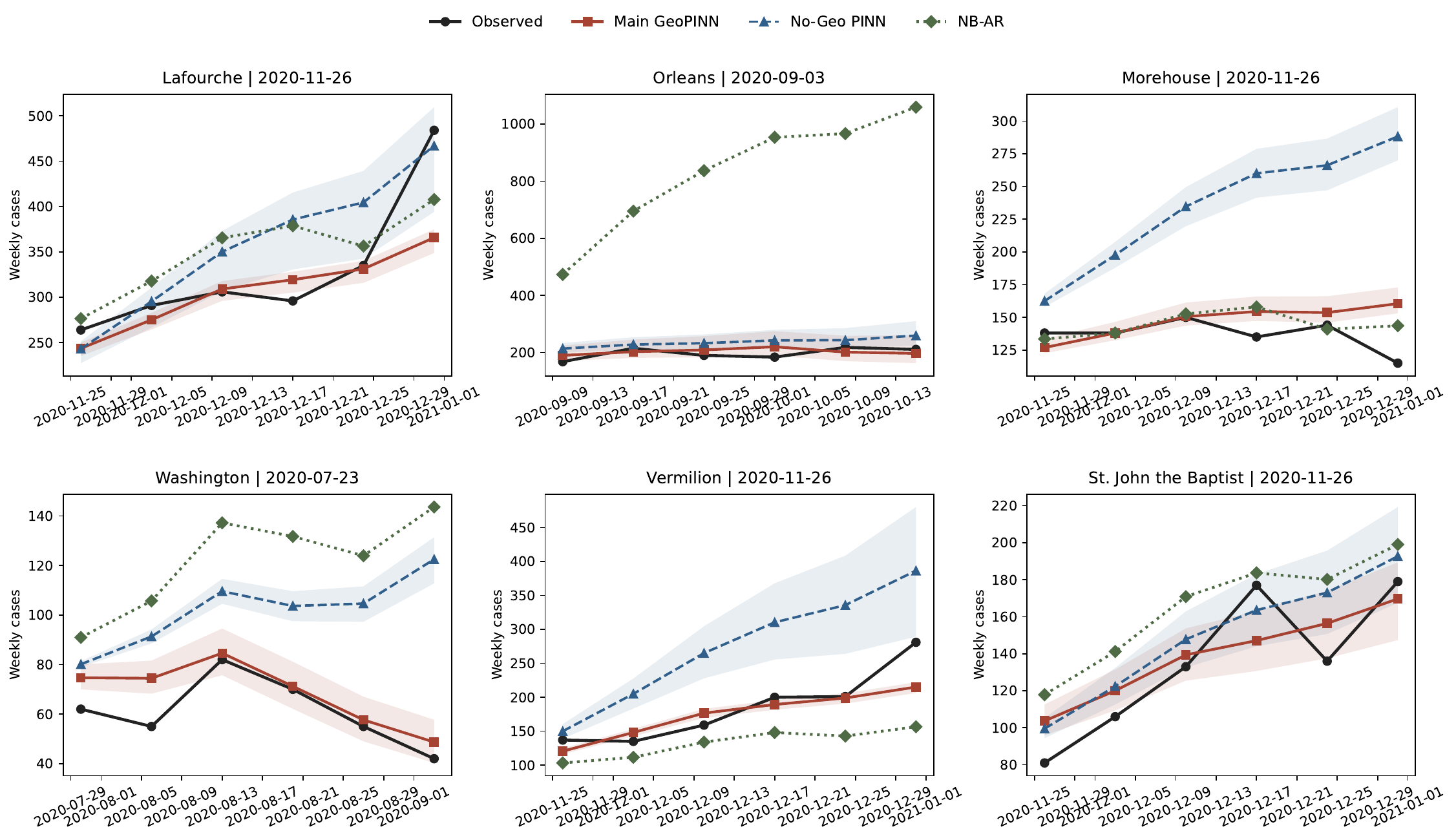}
\caption{Selected six-week Louisiana trajectories. Black circles show observations, red squares Forecast-Trained Geo-PINN, blue triangles the no-geography comparator, and green diamonds fixed-origin NB-AR. Shading spans the three neural seeds for each PINN. The windows were chosen where Forecast-Trained Geo-PINN performed well, so the figure is a conditional trajectory diagnostic rather than an unbiased model comparison.}
\Description{Six panels show observed weekly cases and six-week predictions for selected Louisiana county-origin windows. Black circle lines are observations; red square, blue triangle, and green diamond lines are the Forecast-Trained Geo-PINN, No-Geo PINN, and fixed-origin NB-AR. Shaded bands show the three-seed ranges of the two PINNs.}
\label{fig:selected-window-trajectories}
\end{figure*}

Figure~\ref{fig:selected-window-trajectories} shows six-week trajectories for Forecast-Trained Geo-PINN (labeled Main GeoPINN in the figure), No-Geo PINN (the no-geography comparator), and the fixed-origin Autoregressive Negative-Binomial Baseline (NB-AR) in six selected county--origin windows.

\subsection{Controlled geography and practical-prior robustness}
Table~\ref{tab:louisiana-priors} separates the controlled geography comparison from the 12-prior sensitivity analysis. In Panel A, the Neighbor Network lowers MSE from 31,999 to 29,808, a 6.85\% gain. MAE decreases from 113.26 to 109.80, about 3.1\%, and NLL changes from 6.469 to 6.464. All other inputs and training steps are fixed. This comparison therefore gives the clearest estimate of the incremental value of county geography.

\begin{table*}[tbp]
\caption{Louisiana 15-county high-case subset retrospective geography and prior comparisons; lower MSE, MAE, and NLL are better. Panel A changes only county coupling. Panel B reports the complete spatial-prior sensitivity analysis. Bold marks the lowest value within each panel and metric.}
\label{tab:louisiana-priors}
\centering
\small
\begin{tabularx}{\textwidth}{@{}>{\raggedright\arraybackslash}Xrrr@{}}
\toprule
Model or spatial prior & MSE & MAE & Test NLL \\
\midrule
\multicolumn{4}{@{}l}{\textit{A. Controlled county-network ablation}} \\
No County-Network PINN & 31,999 & 113.26 & 6.469 \\
Neighbor-Network Geo-PINN & \textbf{29,808} & \textbf{109.80} & \textbf{6.464} \\
\addlinespace
\multicolumn{4}{@{}l}{\textit{B. Complete spatial-prior sensitivity}} \\
Data-Inferred Transmission Network & 29,938 & 109.90 & \textbf{6.453} \\
Neighbor Network & \textbf{29,808} & \textbf{109.80} & 6.464 \\
Local-Only Network & 31,999 & 113.26 & 6.469 \\
Mobility-Geography Ensemble Network & 31,789 & 114.53 & 6.480 \\
Bidirectional Commuting Network & 32,774 & 115.44 & 6.484 \\
Neighbor-Distance Network & 31,117 & 114.18 & 6.485 \\
Home-to-Work Commuting Network & 31,751 & 113.74 & 6.485 \\
Distance-Decay Network & 32,329 & 115.41 & 6.488 \\
Work-to-Home Exposure Network & 33,696 & 116.13 & 6.489 \\
Shuffled Commuting Control & 34,746 & 117.22 & 6.493 \\
Farther-County Placebo Network & 37,273 & 121.59 & 6.499 \\
Unregularized Learned Network & 39,721 & 121.46 & 6.549 \\
\bottomrule
\end{tabularx}
\end{table*}

Across the 12 priors, the Neighbor Network has the lowest MSE (29,808) and MAE (109.80). The Data-Inferred Transmission Network is close, with MSE 29,938 and MAE 109.90. It has the lowest mean NLL, 6.453, compared with 6.464 for the Neighbor Network. After false discovery rate (FDR) adjustment, no alternative has a significant NLL difference from the Data-Inferred Network. FDR controls the expected fraction of false discoveries among rejected comparisons. Placebo and unregularized priors perform worse, but several plausible structured priors are nearly equivalent. This pattern supports spatial regularization more strongly than a unique edge-level map.

\section{Discussion and Limitations}
The two studies provide different evidence. Because $\Cmat^*$ is known in the synthetic system, we can measure recovery under controlled priors and observation channels. The main finding is diagnostic: similar trajectory losses can accompany very different matrix errors. Louisiana provides realistic scale and useful baselines, but no edge-level truth. It informs retrospective forecast accuracy and prior robustness, not matrix validation.

Stable $\hat\alpha$ values do not mean that the full inverse problem is well identified. Row normalization makes $\alpha$ a global magnitude parameter, while correlated prevalence leaves source composition weakly constrained. A compatible or misspecified prior can therefore change the matrix without changing the fitted curve much. This is what we expect when the objective is relatively flat in the coupling directions.

The observation process matters as much as the mechanistic residual. In the synthetic study, reporting bias and the cases-only proxy produce large data losses despite small physics losses. Low reporting rates can make a source appear less influential. Reporting delay can create an apparent lead or lag that is unrelated to transmission. The fitted $\Cmat$ may absorb either error, and residual lead--lag priors face the same problem. Epidemiological interpretation therefore requires either an explicit reporting-and-delay model or external calibration. Residuals should then be checked by channel, county, origin, and horizon. Adding flexible reporting, delay, and coupling parameters without new information would worsen, not solve, the identification problem.

The Louisiana results are also retrospective. The external input is observed throughout each test window. An operational system would have to forecast that input and carry its uncertainty into the county forecasts. The network predicts states directly; it does not recursively roll the SIRD equations forward. The 64-county Local-Only comparison also removes the external-input channel, so it bundles several changes. The controlled Neighbor-Network versus No County-Network comparison isolates coupling more cleanly.

Forecast-Trained Geo-PINN lowers MSE from 32,957 to 11,468 and MAE from 70.60 to 57.73 relative to the Autoregressive Negative-Binomial Baseline. The baseline, however, has lower NLL, 5.158 versus 5.346. The metrics answer different questions. MSE and MAE assess the predicted mean. NLL evaluates the probability assigned to the observations under the assumed negative-binomial distribution. The baseline is optimized for that likelihood, whereas Geo-PINN is not; Geo-PINN's NLL uses fixed dispersion $\theta_{\mathrm{NB}}=2$. A lower baseline NLL can therefore coexist with larger point errors. This is a tradeoff, not uniform superiority.

Uncertainty remains limited. Each synthetic grid uses one seed, so multistart confidence intervals are unavailable. Edge-wise profile losses were not completed. The four Louisiana origins and three neural seeds are repeated tests within one setting, not independent external validation. Future work should add multistart summaries, edge-wise profiles or perturbations, independent holdouts, and prospective forecasts of the external input. These checks are necessary before interpreting any row of $\Cmat$ as transmission structure.

The Louisiana data are aggregate county-week counts, not individual records. They support population-level modeling but not person-level inference. Aggregation also does not remove reporting, access, or data-quality bias. We therefore do not claim that the matrix identifies individual contacts, literal mobility, or causal transmission pathways.

Table~\ref{tab:interpretation-rules} summarizes a practical reporting order. The target quantity should be defined before optimization: rows index recipients, columns index sources, and the matrix should be called source composition unless independent flow data support a mobility interpretation. Reports should also state which transmission, removal, reporting, and initial-state quantities are fixed or learned. Every additional free component can enlarge the set of observationally equivalent solutions.

Prior construction should be documented as carefully as the model architecture. A reproducible description gives the raw edge score, direction convention, diagonal mass, row normalization, structural mask, regularization strength, and initialization. A zero penalty is no regularization, even when the same matrix initializes the logits. Sensitivity analyses should include a simple geographic prior, a data-informed prior when available, and a local-only or unregularized reference. Placebo or shuffled scores help show whether any structured prior would perform similarly.

Diagnostics should remain separated by purpose. Report observation losses by channel and mask, physics residuals for all four equations, and coupling sensitivity across priors and initializations. A small total loss can hide a failed observation channel. A small physics loss can coexist with prior-dominated coupling. When feasible, heatmaps should be accompanied by multistart summaries and edge-wise profile or perturbation losses.

Forecast comparisons should separate bundled system changes from controlled geography ablations. MSE and MAE should appear before NLL or other distributional scores, and positive gains should always mean lower point error for the proposed model. Operational claims require every test-period input to be available in real time or forecast separately. This reporting order cannot guarantee identifiability, but it makes assumptions and tradeoffs visible.

We distinguish three claim levels. \emph{Fitted composition} means only that the row-stochastic kernel is part of a model that fits the stated observations. \emph{Predictive geography} requires a controlled ablation in which structured cross-region information improves held-out forecasts. \emph{Recovered coupling} requires known truth or comparably strong evidence, including stability across starts and local edge perturbations. The 15-county Louisiana ablation supports predictive geography, not recovered coupling. The synthetic study tests recovery against $\Cmat^*$, but its one-seed grids limit uncertainty assessment. Neither study establishes prospective deployment because the Louisiana external input is observed during the test window.

Appendix~\ref{app:validation-agenda} expands three next steps: multistart and edge-wise synthetic diagnostics, forecasts of the external input with propagated uncertainty, and tests of structured priors in other jurisdictions.

\FloatBarrier
\section{Conclusion}
Routine surveillance data can support regional epidemic forecasting, but they do not identify cross-region structure on their own. \method addresses this limitation by placing a row-stochastic source-composition matrix inside the SIRD force of infection and regularizing it with empirical spatial information. The model also constrains the neural states to preserve population totals and evaluates observation fit, mechanistic consistency, and matrix recovery separately.

The synthetic study shows why these distinctions matter. A compatible prior improves recovery of the known matrix, while a misspecified prior can change the fitted matrix without materially changing the epidemic trajectory. In Louisiana, county adjacency reduces MSE by 6.85\% and MAE by 3.1\% in the controlled comparison. Forecast-Trained Geo-PINN also reduces MSE by 65.2\% and MAE by 18.2\% relative to the Autoregressive Negative-Binomial Baseline, although the baseline has lower NLL, 5.158 versus 5.346. Several plausible priors perform similarly. We therefore interpret the learned matrix as conditional source composition, not literal mobility or a unique causal network. Future work should model reporting and delay, forecast external inputs prospectively, and measure uncertainty across optimization starts and edge perturbations.

\phantomsection
\label{maincontent:end}

\bibliographystyle{ACM-Reference-Format}
\bibliography{references}


\begin{thebibliography}{22}

%%% the Web address.

\ifx \showCODEN    \undefined \def \showCODEN     #1{\unskip}     \fi
\ifx \showISBNx    \undefined \def \showISBNx     #1{\unskip}     \fi
\ifx \showISBNxiii \undefined \def \showISBNxiii  #1{\unskip}     \fi
\ifx \showISSN     \undefined \def \showISSN      #1{\unskip}     \fi
\ifx \showLCCN     \undefined \def \showLCCN      #1{\unskip}     \fi
\ifx \shownote     \undefined \def \shownote      #1{#1}          \fi
\ifx \showarticletitle \undefined \def \showarticletitle #1{#1}   \fi
\ifx \showURL      \undefined \def \showURL       {\relax}        \fi
% invisible to TeX
\providecommand\bibfield[2]{#2}
\providecommand\bibinfo[2]{#2}
\providecommand\natexlab[1]{#1}
\providecommand\showeprint[2][]{arXiv:#2}

\bibitem[Balcan et~al\mbox{.}(2009)]%
        {balcan2009multiscale}
\bibfield{author}{\bibinfo{person}{Duygu Balcan}, \bibinfo{person}{Vittoria
  Colizza}, \bibinfo{person}{Bruno Gon{\c{c}}alves}, \bibinfo{person}{Hao Hu},
  \bibinfo{person}{Jos{\'e}~J. Ramasco}, {and} \bibinfo{person}{Alessandro
  Vespignani}.} \bibinfo{year}{2009}\natexlab{}.
\newblock \showarticletitle{Multiscale Mobility Networks and the Spatial
  Spreading of Infectious Diseases}.
\newblock \bibinfo{journal}{\emph{Proceedings of the National Academy of
  Sciences of the United States of America}} \bibinfo{volume}{106},
  \bibinfo{number}{51} (\bibinfo{year}{2009}), \bibinfo{pages}{21484--21489}.
\newblock
\href{https://doi.org/10.1073/pnas.0906910106}{doi:\nolinkurl{10.1073/pnas.0906910106}}


\bibitem[Bustamante-Casta{\~n}eda et~al\mbox{.}(2021)]%
        {bustamante2021epidemic}
\bibfield{author}{\bibinfo{person}{F. Bustamante-Casta{\~n}eda},
  \bibinfo{person}{Jean-Guy Caputo}, \bibinfo{person}{Gustavo Cruz-Pacheco},
  \bibinfo{person}{Arnaud Knippel}, {and} \bibinfo{person}{Fatima Mouatamide}.}
  \bibinfo{year}{2021}\natexlab{}.
\newblock \showarticletitle{Epidemic Model on a Network: Analysis and
  Applications to {COVID-19}}.
\newblock \bibinfo{journal}{\emph{Physica A: Statistical Mechanics and its
  Applications}}  \bibinfo{volume}{564} (\bibinfo{year}{2021}),
  \bibinfo{pages}{125520}.
\newblock
\href{https://doi.org/10.1016/j.physa.2020.125520}{doi:\nolinkurl{10.1016/j.physa.2020.125520}}


\bibitem[Colizza et~al\mbox{.}(2006)]%
        {colizza2006airline}
\bibfield{author}{\bibinfo{person}{Vittoria Colizza}, \bibinfo{person}{Alain
  Barrat}, \bibinfo{person}{Marc Barth{\'e}lemy}, {and}
  \bibinfo{person}{Alessandro Vespignani}.} \bibinfo{year}{2006}\natexlab{}.
\newblock \showarticletitle{The Role of the Airline Transportation Network in
  the Prediction and Predictability of Global Epidemics}.
\newblock \bibinfo{journal}{\emph{Proceedings of the National Academy of
  Sciences of the United States of America}} \bibinfo{volume}{103},
  \bibinfo{number}{7} (\bibinfo{year}{2006}), \bibinfo{pages}{2015--2020}.
\newblock
\href{https://doi.org/10.1073/pnas.0510525103}{doi:\nolinkurl{10.1073/pnas.0510525103}}


\bibitem[Diekmann et~al\mbox{.}(2010)]%
        {diekmann2010nextgeneration}
\bibfield{author}{\bibinfo{person}{Odo Diekmann}, \bibinfo{person}{J.~A.~P.
  Heesterbeek}, {and} \bibinfo{person}{Michael~G. Roberts}.}
  \bibinfo{year}{2010}\natexlab{}.
\newblock \showarticletitle{The Construction of Next-Generation Matrices for
  Compartmental Epidemic Models}.
\newblock \bibinfo{journal}{\emph{Journal of the Royal Society Interface}}
  \bibinfo{volume}{7}, \bibinfo{number}{47} (\bibinfo{year}{2010}),
  \bibinfo{pages}{873--885}.
\newblock
\href{https://doi.org/10.1098/rsif.2009.0386}{doi:\nolinkurl{10.1098/rsif.2009.0386}}


\bibitem[Eubank et~al\mbox{.}(2004)]%
        {eubank2004modeling}
\bibfield{author}{\bibinfo{person}{Stephen Eubank}, \bibinfo{person}{Hasan
  Guclu}, \bibinfo{person}{V.~S.~Anil Kumar}, \bibinfo{person}{Madhav~V.
  Marathe}, \bibinfo{person}{Aravind Srinivasan}, \bibinfo{person}{Zoltan
  Toroczkai}, {and} \bibinfo{person}{Nan Wang}.}
  \bibinfo{year}{2004}\natexlab{}.
\newblock \showarticletitle{Modelling Disease Outbreaks in Realistic Urban
  Social Networks}.
\newblock \bibinfo{journal}{\emph{Nature}} \bibinfo{volume}{429},
  \bibinfo{number}{6988} (\bibinfo{year}{2004}), \bibinfo{pages}{180--184}.
\newblock
\href{https://doi.org/10.1038/nature02541}{doi:\nolinkurl{10.1038/nature02541}}


\bibitem[Finkenst{\"a}dt et~al\mbox{.}(2002)]%
        {finkenstadt2002stochastic}
\bibfield{author}{\bibinfo{person}{B{\"a}rbel~F. Finkenst{\"a}dt},
  \bibinfo{person}{Ottar~N. Bj{\o}rnstad}, {and} \bibinfo{person}{Bryan~T.
  Grenfell}.} \bibinfo{year}{2002}\natexlab{}.
\newblock \showarticletitle{A Stochastic Model for Extinction and Recurrence of
  Epidemics: Estimation and Inference for Measles Outbreaks}.
\newblock \bibinfo{journal}{\emph{Biostatistics}} \bibinfo{volume}{3},
  \bibinfo{number}{4} (\bibinfo{year}{2002}), \bibinfo{pages}{493--510}.
\newblock
\href{https://doi.org/10.1093/biostatistics/3.4.493}{doi:\nolinkurl{10.1093/biostatistics/3.4.493}}


\bibitem[Finkenst{\"a}dt and Grenfell(1998)]%
        {finkenstadt1998empirical}
\bibfield{author}{\bibinfo{person}{B{\"a}rbel~F. Finkenst{\"a}dt} {and}
  \bibinfo{person}{Bryan~T. Grenfell}.} \bibinfo{year}{1998}\natexlab{}.
\newblock \showarticletitle{Empirical Determinants of Measles Metapopulation
  Dynamics in England and Wales}.
\newblock \bibinfo{journal}{\emph{Proceedings of the Royal Society B:
  Biological Sciences}} \bibinfo{volume}{265}, \bibinfo{number}{1392}
  (\bibinfo{year}{1998}), \bibinfo{pages}{211--220}.
\newblock
\href{https://doi.org/10.1098/rspb.1998.0284}{doi:\nolinkurl{10.1098/rspb.1998.0284}}


\bibitem[Hethcote(2000)]%
        {hethcote2000mathematics}
\bibfield{author}{\bibinfo{person}{Herbert~W. Hethcote}.}
  \bibinfo{year}{2000}\natexlab{}.
\newblock \showarticletitle{The Mathematics of Infectious Diseases}.
\newblock \bibinfo{journal}{\emph{SIAM Rev.}} \bibinfo{volume}{42},
  \bibinfo{number}{4} (\bibinfo{year}{2000}), \bibinfo{pages}{599--653}.
\newblock
\href{https://doi.org/10.1137/S0036144500371907}{doi:\nolinkurl{10.1137/S0036144500371907}}


\bibitem[Hu et~al\mbox{.}(2022)]%
        {hu2022modified}
\bibfield{author}{\bibinfo{person}{Haoran Hu}, \bibinfo{person}{Connor~M.
  Kennedy}, \bibinfo{person}{Panayotis~G. Kevrekidis}, {and}
  \bibinfo{person}{Hong-Kun Zhang}.} \bibinfo{year}{2022}\natexlab{}.
\newblock \showarticletitle{A Modified {PINN} Approach for Identifiable
  Compartmental Models in Epidemiology with Application to {COVID-19}}.
\newblock \bibinfo{journal}{\emph{Viruses}} \bibinfo{volume}{14},
  \bibinfo{number}{11} (\bibinfo{year}{2022}), \bibinfo{pages}{2464}.
\newblock
\href{https://doi.org/10.3390/v14112464}{doi:\nolinkurl{10.3390/v14112464}}


\bibitem[Karniadakis et~al\mbox{.}(2021)]%
        {karniadakis2021piml}
\bibfield{author}{\bibinfo{person}{George~Em Karniadakis},
  \bibinfo{person}{Ioannis~G. Kevrekidis}, \bibinfo{person}{Lu Lu},
  \bibinfo{person}{Paris Perdikaris}, \bibinfo{person}{Sifan Wang}, {and}
  \bibinfo{person}{Liu Yang}.} \bibinfo{year}{2021}\natexlab{}.
\newblock \showarticletitle{Physics-Informed Machine Learning}.
\newblock \bibinfo{journal}{\emph{Nature Reviews Physics}} \bibinfo{volume}{3},
  \bibinfo{number}{6} (\bibinfo{year}{2021}), \bibinfo{pages}{422--440}.
\newblock
\href{https://doi.org/10.1038/s42254-021-00314-5}{doi:\nolinkurl{10.1038/s42254-021-00314-5}}


\bibitem[Keeling and Eames(2005)]%
        {keeling2005networks}
\bibfield{author}{\bibinfo{person}{Matt~J. Keeling} {and} \bibinfo{person}{Ken
  T.~D. Eames}.} \bibinfo{year}{2005}\natexlab{}.
\newblock \showarticletitle{Networks and Epidemic Models}.
\newblock \bibinfo{journal}{\emph{Journal of the Royal Society Interface}}
  \bibinfo{volume}{2}, \bibinfo{number}{4} (\bibinfo{year}{2005}),
  \bibinfo{pages}{295--307}.
\newblock
\href{https://doi.org/10.1098/rsif.2005.0051}{doi:\nolinkurl{10.1098/rsif.2005.0051}}


\bibitem[Kermack and McKendrick(1927)]%
        {kermack1927contribution}
\bibfield{author}{\bibinfo{person}{William~Ogilvy Kermack} {and}
  \bibinfo{person}{Anderson~Gray McKendrick}.} \bibinfo{year}{1927}\natexlab{}.
\newblock \showarticletitle{A Contribution to the Mathematical Theory of
  Epidemics}.
\newblock \bibinfo{journal}{\emph{Proceedings of the Royal Society of London.
  Series A, Containing Papers of a Mathematical and Physical Character}}
  \bibinfo{volume}{115}, \bibinfo{number}{772} (\bibinfo{year}{1927}),
  \bibinfo{pages}{700--721}.
\newblock
\href{https://doi.org/10.1098/rspa.1927.0118}{doi:\nolinkurl{10.1098/rspa.1927.0118}}


\bibitem[Kharazmi et~al\mbox{.}(2021)]%
        {kharazmi2021pinn}
\bibfield{author}{\bibinfo{person}{Ehsan Kharazmi}, \bibinfo{person}{Min Cai},
  \bibinfo{person}{Xiaoning Zheng}, \bibinfo{person}{Zhen Zhang},
  \bibinfo{person}{Guang Lin}, {and} \bibinfo{person}{George~Em Karniadakis}.}
  \bibinfo{year}{2021}\natexlab{}.
\newblock \showarticletitle{Identifiability and Predictability of Integer- and
  Fractional-Order Epidemiological Models Using Physics-Informed Neural
  Networks}.
\newblock \bibinfo{journal}{\emph{Nature Computational Science}}
  \bibinfo{volume}{1}, \bibinfo{number}{11} (\bibinfo{year}{2021}),
  \bibinfo{pages}{744--753}.
\newblock
\href{https://doi.org/10.1038/s43588-021-00158-0}{doi:\nolinkurl{10.1038/s43588-021-00158-0}}


\bibitem[Massonis et~al\mbox{.}(2021)]%
        {massonis2021structural}
\bibfield{author}{\bibinfo{person}{Gemma Massonis}, \bibinfo{person}{Julio~R.
  Banga}, {and} \bibinfo{person}{Alejandro~F. Villaverde}.}
  \bibinfo{year}{2021}\natexlab{}.
\newblock \showarticletitle{Structural Identifiability and Observability of
  Compartmental Models of the {COVID-19} Pandemic}.
\newblock \bibinfo{journal}{\emph{Annual Reviews in Control}}
  \bibinfo{volume}{51} (\bibinfo{year}{2021}), \bibinfo{pages}{441--459}.
\newblock
\href{https://doi.org/10.1016/j.arcontrol.2020.12.001}{doi:\nolinkurl{10.1016/j.arcontrol.2020.12.001}}


\bibitem[Millevoi et~al\mbox{.}(2024)]%
        {millevoi2024pinn}
\bibfield{author}{\bibinfo{person}{Caterina Millevoi}, \bibinfo{person}{Damiano
  Pasetto}, {and} \bibinfo{person}{Massimiliano Ferronato}.}
  \bibinfo{year}{2024}\natexlab{}.
\newblock \showarticletitle{A Physics-Informed Neural Network Approach for
  Compartmental Epidemiological Models}.
\newblock \bibinfo{journal}{\emph{PLOS Computational Biology}}
  \bibinfo{volume}{20}, \bibinfo{number}{9} (\bibinfo{year}{2024}),
  \bibinfo{pages}{e1012387}.
\newblock
\href{https://doi.org/10.1371/journal.pcbi.1012387}{doi:\nolinkurl{10.1371/journal.pcbi.1012387}}


\bibitem[Noppert et~al\mbox{.}(2024)]%
        {noppert2024state}
\bibfield{author}{\bibinfo{person}{Grace~A. Noppert}, \bibinfo{person}{Philippa
  Clarke}, \bibinfo{person}{Andrew Hoover}, \bibinfo{person}{John Kubale},
  \bibinfo{person}{Robert Melendez}, \bibinfo{person}{Kate Duchowny}, {and}
  \bibinfo{person}{Sonia~T. Hegde}.} \bibinfo{year}{2024}\natexlab{}.
\newblock \showarticletitle{State Variation in Neighborhood {COVID-19} Burden
  across the United States}.
\newblock \bibinfo{journal}{\emph{Communications Medicine}}
  \bibinfo{volume}{4} (\bibinfo{year}{2024}), \bibinfo{pages}{36}.
\newblock
\href{https://doi.org/10.1038/s43856-024-00459-1}{doi:\nolinkurl{10.1038/s43856-024-00459-1}}


\bibitem[Pei et~al\mbox{.}(2018)]%
        {pei2018forecasting}
\bibfield{author}{\bibinfo{person}{Sen Pei}, \bibinfo{person}{Sasikiran
  Kandula}, \bibinfo{person}{Wan Yang}, {and} \bibinfo{person}{Jeffrey
  Shaman}.} \bibinfo{year}{2018}\natexlab{}.
\newblock \showarticletitle{Forecasting the Spatial Transmission of Influenza
  in the United States}.
\newblock \bibinfo{journal}{\emph{Proceedings of the National Academy of
  Sciences of the United States of America}} \bibinfo{volume}{115},
  \bibinfo{number}{11} (\bibinfo{year}{2018}), \bibinfo{pages}{2752--2757}.
\newblock
\href{https://doi.org/10.1073/pnas.1708856115}{doi:\nolinkurl{10.1073/pnas.1708856115}}


\bibitem[Raissi et~al\mbox{.}(2019)]%
        {raissi2019pinn}
\bibfield{author}{\bibinfo{person}{Maziar Raissi}, \bibinfo{person}{Paris
  Perdikaris}, {and} \bibinfo{person}{George~Em Karniadakis}.}
  \bibinfo{year}{2019}\natexlab{}.
\newblock \showarticletitle{Physics-Informed Neural Networks: A Deep Learning
  Framework for Solving Forward and Inverse Problems Involving Nonlinear
  Partial Differential Equations}.
\newblock \bibinfo{journal}{\emph{J. Comput. Phys.}}  \bibinfo{volume}{378}
  (\bibinfo{year}{2019}), \bibinfo{pages}{686--707}.
\newblock
\href{https://doi.org/10.1016/j.jcp.2018.10.045}{doi:\nolinkurl{10.1016/j.jcp.2018.10.045}}


\bibitem[Tuncer and Le(2018)]%
        {tuncer2018identifiability}
\bibfield{author}{\bibinfo{person}{Necibe Tuncer} {and}
  \bibinfo{person}{Trang~T. Le}.} \bibinfo{year}{2018}\natexlab{}.
\newblock \showarticletitle{Structural and Practical Identifiability Analysis
  of Outbreak Models}.
\newblock \bibinfo{journal}{\emph{Mathematical Biosciences}}
  \bibinfo{volume}{299} (\bibinfo{year}{2018}), \bibinfo{pages}{1--18}.
\newblock
\href{https://doi.org/10.1016/j.mbs.2018.02.004}{doi:\nolinkurl{10.1016/j.mbs.2018.02.004}}


\bibitem[Viboud et~al\mbox{.}(2006)]%
        {viboud2006synchrony}
\bibfield{author}{\bibinfo{person}{C{\'e}cile Viboud},
  \bibinfo{person}{Ottar~N. Bj{\o}rnstad}, \bibinfo{person}{David~L. Smith},
  \bibinfo{person}{Lone Simonsen}, \bibinfo{person}{Mark~A. Miller}, {and}
  \bibinfo{person}{Bryan~T. Grenfell}.} \bibinfo{year}{2006}\natexlab{}.
\newblock \showarticletitle{Synchrony, Waves, and Spatial Hierarchies in the
  Spread of Influenza}.
\newblock \bibinfo{journal}{\emph{Science}} \bibinfo{volume}{312},
  \bibinfo{number}{5772} (\bibinfo{year}{2006}), \bibinfo{pages}{447--451}.
\newblock
\href{https://doi.org/10.1126/science.1125237}{doi:\nolinkurl{10.1126/science.1125237}}


\bibitem[Wang et~al\mbox{.}(2021)]%
        {wang2021gradient}
\bibfield{author}{\bibinfo{person}{Sifan Wang}, \bibinfo{person}{Yujun Teng},
  {and} \bibinfo{person}{Paris Perdikaris}.} \bibinfo{year}{2021}\natexlab{}.
\newblock \showarticletitle{Understanding and Mitigating Gradient Flow
  Pathologies in Physics-Informed Neural Networks}.
\newblock \bibinfo{journal}{\emph{SIAM Journal on Scientific Computing}}
  \bibinfo{volume}{43}, \bibinfo{number}{5} (\bibinfo{year}{2021}),
  \bibinfo{pages}{A3055--A3081}.
\newblock
\href{https://doi.org/10.1137/20M1318043}{doi:\nolinkurl{10.1137/20M1318043}}


\bibitem[Xia et~al\mbox{.}(2004)]%
        {xia2004measles}
\bibfield{author}{\bibinfo{person}{Yingcun Xia}, \bibinfo{person}{Ottar~N.
  Bj{\o}rnstad}, {and} \bibinfo{person}{Bryan~T. Grenfell}.}
  \bibinfo{year}{2004}\natexlab{}.
\newblock \showarticletitle{Measles Metapopulation Dynamics: A Gravity Model
  for Epidemiological Coupling and Dynamics}.
\newblock \bibinfo{journal}{\emph{The American Naturalist}}
  \bibinfo{volume}{164}, \bibinfo{number}{2} (\bibinfo{year}{2004}),
  \bibinfo{pages}{267--281}.
\newblock
\href{https://doi.org/10.1086/422341}{doi:\nolinkurl{10.1086/422341}}


\end{thebibliography}

\appendix
\section{Exploratory Policy-Stratified Geography Results}
\label{app:policy-stratified}
The supplied policy phase is a categorical label for each forecast origin, and the restriction index is the supplied integer policy category. The top-30\% MSE-gain count is the supplied number of county windows assigned to that relative-gain category. An em dash in the NLL column denotes an unavailable value. These fields are descriptive and are not treated as causal exposures.
\begin{table*}[t]
\caption{Exploratory policy-stratified results for the controlled 15-county comparison. Positive MSE gain favors the Neighbor-Network Geo-PINN; the joint column counts windows with gains in both MSE and MAE. NLL was not supplied.}
\label{tab:policy-stratified}
\centering
\small
\begin{tabularx}{\textwidth}{@{}>{\raggedright\arraybackslash}p{0.23\textwidth}r>{\centering\arraybackslash}p{0.20\textwidth}c>{\centering\arraybackslash}p{0.11\textwidth}>{\centering\arraybackslash}X@{}}
\toprule
Forecast condition & Mean relative MSE gain (\%) & MAE: county windows with both gains & NLL & Restriction index & Top-30\% MSE-gain count \\
\midrule
July 23 / Phase 2 & $+3.61$ & 9/15 (60.0\%) & --- & 3 & 4 \\
September 3 / Phase 2 to Phase 3 & $-11.49$ & 6/15 (40.0\%) & --- & 2 & 3 \\
October 15 / Phase 3 & $+2.65$ & 11/15 (73.3\%) & --- & 1 & 6 \\
November 26 / Modified Phase 2 & $+2.55$ & 7/15 (46.7\%) & --- & 4 & 5 \\
\bottomrule
\end{tabularx}
\end{table*}

Within the controlled 15-county comparison, Neighbor-Network Geo-PINN is the proposed model. No County-Network PINN is the baseline. A \emph{county window} is one county at one forecast origin. Mean relative MSE gains across the four origins are $+3.61\%$, $-11.49\%$, $+2.65\%$, and $+2.55\%$. Both MSE and MAE improve in 9/15 (60.0\%), 6/15 (40.0\%), 11/15 (73.3\%), and 7/15 (46.7\%) county windows. Origin-specific NLL was not supplied, so the table uses an em dash. The unfavorable September MSE result remains visible without an unsupported distributional claim.

Table~\ref{tab:policy-stratified} reports relative MSE gain first, then the count and proportion of windows improving on both MSE and MAE, followed by NLL and the remaining fields. The analysis is post hoc and descriptive. The sign convention follows Eq.~\eqref{eq:relative-gains}. Policy phase and restriction index are supplied categorical labels, not causal exposures. The final column reproduces the supplied top-30\% MSE-gain count; its ranking reference set still requires author verification.

\section{Terminology and Scope Crosswalk}
\label{app:crosswalk}
The paper uses the following names consistently. Forecast-Trained Geo-PINN is the final 64-county model. Data-Inferred Transmission Network is the adaptive residual lead--lag prior. Neighbor Network is county adjacency, and Local-Only Network is the identity prior. The controlled pair consists of the No County-Network PINN with identity coupling and the Neighbor-Network Geo-PINN with adjacency. Autoregressive Negative-Binomial Baseline is the negative-binomial autoregressive comparator.

The five observation-stress settings are clean full-state, noisy full-state, weekly sparse, unmodeled reporting bias, and a cases-only proxy. We do not claim any additional noise distribution, confidence interval, profile loss, or larger synthetic benchmark.

\section{Secondary Synthetic Prior Quantities}
\label{app:secondary-synthetic}
The absolute transmission-scale errors $|\hat\alpha-\alpha^*|$ are 0.043, 0.043, 0.043, 0.042, 0.045, and 0.048 for Distance weak, Distance medium, No prior, Too diagonal, Too mixed, and Wrong strong. These small differences reinforce the main result: a stable transmission magnitude does not imply a stable geographic composition.

\section{Interpretation Rules}
\label{app:interpretation-rules}
\begin{table}[H]
\caption{Interpretation rules and evidence in this paper.}
\label{tab:interpretation-rules}
\centering
\small
\begin{tabularx}{\columnwidth}{@{}>{\raggedright\arraybackslash}p{0.29\columnwidth}X@{}}
\toprule
Claim & Required diagnostic and status in this paper \\
\midrule
Observation fit & Channel-specific residuals or held-out metrics; biased and cases-only synthetic channels fail despite low physics loss. \\
Mechanistic consistency & All four normalized residuals; a small value does not validate the observation model. \\
Coupling recovery & Known truth or multistart and edge-wise profiles; known truth exists only in the one-seed benchmark. \\
Incremental geography value & Matched models differing only in coupling; provided by the 15-county controlled ablation. \\
Operational forecast value & Prospective inputs, independent holdout, and calibrated distributions; not established here. \\
\bottomrule
\end{tabularx}
\end{table}

\section{Expanded Validation Agenda}
\label{app:validation-agenda}
Future synthetic work should vary the number of regions, correlation among prevalence curves, seeding, observation channels, and prior misspecification. Each setting should use several neural and coupling initializations. Multistart summaries of $\hat\alpha$, matrix error, row-wise error, and observation loss, together with edge-wise profiles or perturbation losses, would show whether coupling entries are constrained by the data or selected mainly by the prior. These outputs are unavailable here, so the one-seed grids remain diagnostic.

Real-data work should replace the oracle input with forecasts made at each origin and propagate their uncertainty. Negative-binomial dispersion should be selected within training data or replaced by a justified observation model, without tuning on the test windows. Repeating the rolling-origin analysis in other jurisdictions would test generalizability and should preserve the distinction between direct neural prediction and recursive rollout.

Adjacency, distance, commuting, and data-inferred scores can support triangulation without serving as ground truth. Agreement would strengthen a qualitative claim; similar forecast performance would indicate weak edge identification. More complex kernels, delay models, or movement operators require additional observations and regularization. The goal is the strongest claim that remains stable across alternatives, not the most detailed fitted matrix.

\end{document}